%% file: main.tex
\documentclass[10pt,twocolumn,letterpaper]{article}

\usepackage{cvpr}              

\usepackage{graphicx}
\usepackage{amsmath}
\usepackage{amssymb}
\usepackage{booktabs}
\usepackage{multirow}
\usepackage{bm}
\usepackage{enumitem}
\usepackage{xcolor}

\usepackage[pagebackref=false, breaklinks=true, colorlinks, citecolor=citecolor, linkcolor=linkcolor, bookmarks=false]{hyperref}
\definecolor{citecolor}{HTML}{0071BC}
\definecolor{linkcolor}{HTML}{ED1C24}

\usepackage[capitalize]{cleveref}
\crefname{section}{Sec.}{Secs.}
\Crefname{section}{Section}{Sections}
\Crefname{table}{Table}{Tables}
\crefname{table}{Tab.}{Tabs.}

\def\cvprPaperID{2296} 
\def\confName{CVPR}
\def\confYear{2022}

\newcommand{\x}{\bm{x}}
\newcommand{\y}{\bm{y}}
\newcommand{\z}{\bm{z}}

\renewcommand{\medskip}{\vspace{0.02in}}

\usepackage{xcolor,colortbl}
\definecolor{Gray}{gray}{0.9}

\begin{document}

\title{Learning Where to Learn in Cross-View Self-Supervised Learning}

\author{
    Lang Huang$^1$, Shan You$^{2}$\thanks{Corresponding author.\quad $^\dag$Code: \url{https://t.ly/ZI0A}.},~~Mingkai Zheng$^3$, Fei Wang$^2$, Chen Qian$^2$, Toshihiko Yamasaki$^1$\\
    $^1$The University of Tokyo;
    $^2$SenseTime Research;
    $^3$The University of Sydney \\
    {\tt \small \{langhuang,yamasaki\}@cvm.t.u-tokyo.ac.jp}\\ 
    {\tt \small \{youshan,wangfei,qianchen\}@sensetime.com},
    {\tt\small mzhe4001@uni.sydney.edu.au}}

\maketitle

\input{chaps/abstract}

\input{chaps/intro}
\input{chaps/related}
\input{chaps/approach}
\input{chaps/exp}
\input{chaps/conclusion}

{\small
\bibliographystyle{ieee_fullname}
\bibliography{main}
}

\input{chaps/appendix}

\end{document}

%% file: chaps/abstract.tex
\begin{abstract}

Self-supervised learning (SSL) has made enormous progress and largely narrowed the gap with the supervised ones, where the representation learning is mainly guided by a projection into an embedding space. During the projection, current methods simply adopt uniform aggregation of pixels for embedding; however, this risks involving object-irrelevant nuisances and spatial misalignment for different augmentations. In this paper, we present a new approach, \textbf{Le}arning \textbf{W}her\textbf{e} to \textbf{L}earn (LEWEL), to adaptively aggregate spatial information of features, so that the projected embeddings could be exactly aligned and thus guide the feature learning better. Concretely, we reinterpret the projection head in SSL as a per-pixel projection and predict a set of spatial alignment maps from the original features by this weight-sharing projection head. A spectrum of aligned embeddings is thus obtained by aggregating the features with spatial weighting according to these alignment maps. As a result of this adaptive alignment, we observe substantial improvements on both image-level prediction and dense prediction at the same time: LEWEL improves MoCov2~\cite{he2020momentum} by 1.6\%/1.3\%/0.5\%/0.4\% points, improves BYOL~\cite{grill2020bootstrap} by 1.3\%/1.3\%/0.7\%/0.6\% points, on ImageNet linear/semi-supervised classification, Pascal VOC semantic segmentation, and object detection, respectively.$^\dag$

\end{abstract}

%% file: chaps/intro.tex
\section{Introduction}
\label{sec:intro}

In recent years, self-supervised learning (SSL)~\cite{noroozi2016unsupervised,gidaris2018unsupervised,wu2018unsupervised,oord2018representation,he2020momentum,chen2020simple,grill2020bootstrap} has attained tremendous attention due to its impressive ability to learn good representations from large volume of unlabeled data. Among them, the state-of-the-art instance discrimination approaches~\cite{wu2018unsupervised,oord2018representation,he2020momentum,chen2020simple,grill2020bootstrap} encourage the representation learning with image-level invariance to a set of random data transformations, \eg, random cropping and color distortions. These methods even exhibit superior performance over their supervised counterparts for various downstream tasks, such as object detection~\cite{everingham15,lin2014microsoft} and semantic segmentation~\cite{everingham15}. There remain, however, several important issues unresolved. Two of them are mainly attributed to the rigorous invariance to random cropping because it would risk introducing more irrelevant nuisance (\eg, background information) and spatial misalignment of objects for augmentations.
Though, for SSL, random cropping might be the most effective data augmentation option~\cite{chen2020simple} and a good degree of spatial misalignment is beneficial~\cite{tian2020makes}, it remains unclear how to choose the optimal degree of misalignment. Furthermore, the involved nuisance will hinder the discrimination ability of image-level representations while the misalignment discards some important spatial information of objects.

Several recent literatures have dedicated to alleviating these issues by involving some localization priors of downstream tasks in advance. For example, the works of~\cite{wang2021dense,xie2021propagate} explored pixel-level consistency between two augmented views, while some other works proposed to match the representation of a set of pre-defined bounding-boxes~\cite{roh2021spatially,xiao2021region} or pre-computed masks~\cite{henaff2021efficient} between the two views. Despite the improved performance on dense prediction tasks, these methods still suffer from several drawbacks, \eg, they rely on the prior from a specific downstream task and fail to generalize to other tasks. Specifically, there is an undesirable trend that these methods perform worse on the classification task than their instance discrimination counterparts, since they are 
delicatedly tailored for dense predictions and emphasis on the local feature learning.

\begin{figure}
    \centering
    \includegraphics[width=\linewidth]{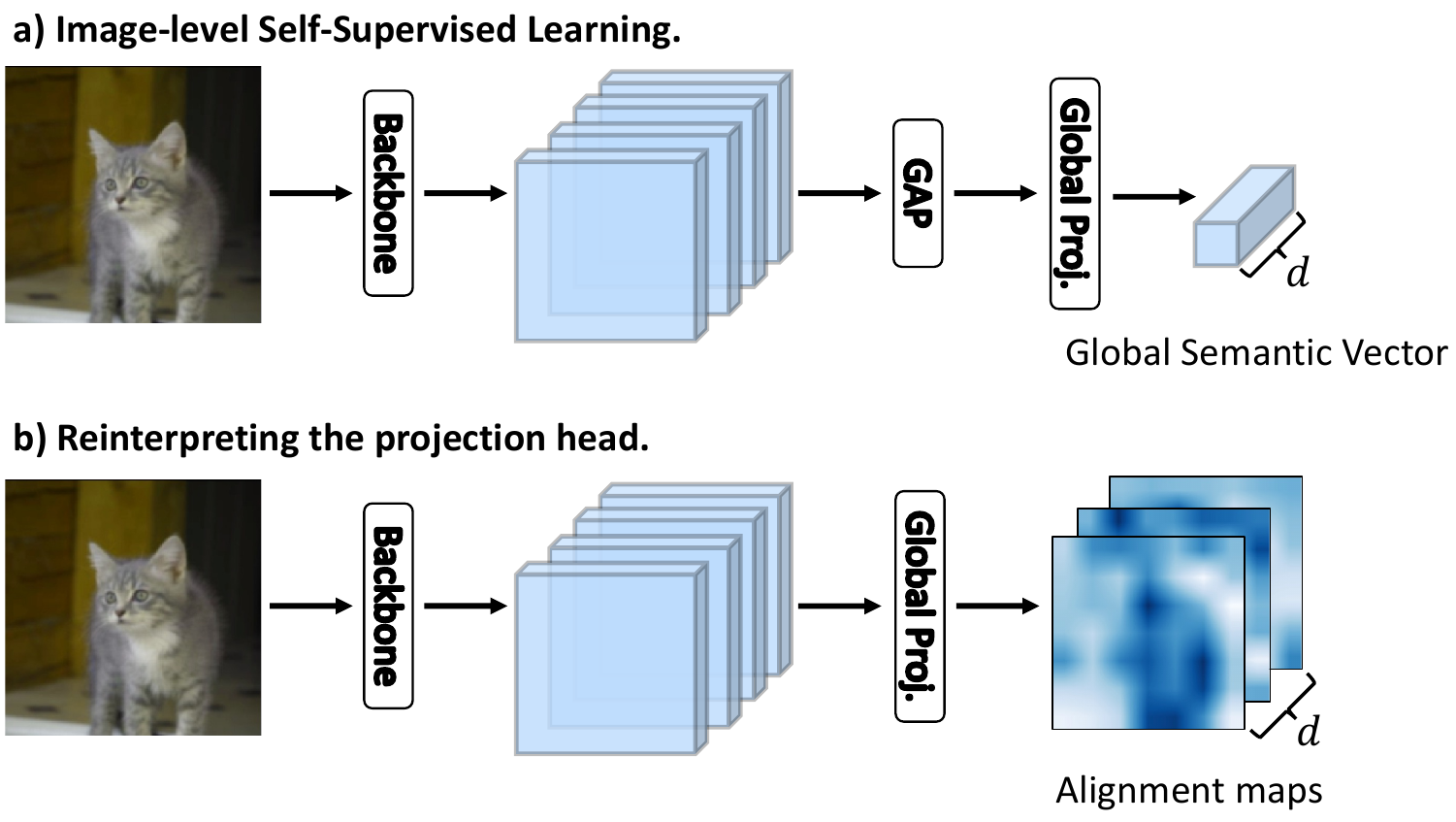}
    \vskip -0.1in
    \caption{
    \textbf{An illustration of the alignment maps prediction}. Based on the standard image-level self-supervised learning, we remove the global average pooling (GAP) layer and reinterpret the global projection as a per-pixel projection, yielding a set of alignment maps that automatically activate to a certain semantic.
    }
    \label{fig:reinterpret}
    \vspace{-0.2in}
\end{figure}

We argue that a good self-supervised representation learning algorithm should not leverage task-specific priors but learn local representations spontaneously. In this paper, we present a new self-supervised learning approach, \textbf{Le}arning \textbf{W}her\textbf{e} to \textbf{L}earn (LEWEL) in a pure end-to-end manner. We first regard the spatial aggregation for embeddings of existing SSL methods, \eg, by the global average pooling (GAP), as summations over all spatial pixels weighted by a set of alignment maps. This formulation suggests that we can explicitly control where to learn in SSL by manipulating the alignment maps. Moreover, in contrast to previous works
manually specifying alignment maps based on downstream rules, our approach learns to just predict the alignment maps on-the-fly during training. 

Nevertheless, it is rather challenging to directly model the alignment maps without any supervised signal. Therefore, we propose a novel reinterpretation scheme to guide this process with the help of the global representation.
As shown in \cref{fig:reinterpret}b, rather than using additional parametrization, we reinterpret the global projection head in SSL as a per-pixel projection to directly predict these alignment maps, which is inspired by the learning paradigm of semantic segmentation~\cite{long2015fully}. In this way, the global embeddings and the semantic alignment maps are coupled with the weight-sharing projection head. This reinterpretation enables the models to automatically find semantically consistent alignment, yielding a set of spatial alignment maps to ``supervise" the alignment process. Actually, the resultant coupled projection head in LEWEL has multiple advantages. On the one hand, based on the generated alignment maps, we can obtain a spectrum of aligned embeddings. Implementing SSL with these embeddings is thus expected to benefit the representation learning since irrelevant nuisances and spatial misalignment have been resolved to a great extent.  On the other hand, learning with the aligned embeddings in return facilitates the global representation to extract more truly discriminative features. As a result of this adaptive alignment and coupled projection head, we observe significantly improvements on both image-level and dense predictions simultaneously. Our main contributions are summarized as follows:

\vspace{-0.1in}
\begin{enumerate}[leftmargin=1.3em,itemsep=0.015em]
    \item We introduce a new self-supervised learning approach, \textbf{Le}arning \textbf{W}her\textbf{e} to \textbf{L}earn (LEWEL), which is a general end-to-end framework and does not involve any downstream task prior for more adaptive and accurate representation learning.
    
    \item In LEWEL, we propose a novel reinterpretation scheme to generate alignment maps with coupled projection head, thus the aligned and global embeddings can be reciprocal to each other, and, as a result, boost the learned representations.
    
    \item LEWEL brings substantial improvements over prior arts on both image-level prediction and dense prediction. We perform extensive evaluation on linear/semi-supervised classification, semantic segmentation, and object detection tasks, using ImageNet-1K~\cite{russakovsky2015imagenet}, Pascal VOC~\cite{everingham15}, and MS-COCO~\cite{lin2014microsoft} benchmarks. Experimental results suggest that LEWEL is able to improve the strong baselines MoCov2~\cite{he2020momentum} and BYOL~\cite{grill2020bootstrap} under all settings. Specifically, LEWEL improves MoCov2~\cite{he2020momentum} by 1.6\%/1.3\%/0.5\%/0.4\% points, improves BYOL~\cite{grill2020bootstrap} by 1.3\%/1.3\%/0.7\%/0.6\% points, on ImageNet linear/semi-supervised classification, Pascal VOC semantic segmentation, and object detection, respectively.
\end{enumerate}

%% file: chaps/related.tex
\section{Related Works}
\label{sec:related}
Learning good representations has long been one of the fundamental questions in computer vision. In recent years, self-supervised learning (SSL) emerged as a promising learning paradigm for representation learning. In general, SSL approaches solve a proxy task to drive the training, including recovering input using auto-encoder~\cite{vincent2008extracting,pathak2016context}, generating pixels in the input space~\cite{kingma2014auto,goodfellow2014generative}, predicting rotation~\cite{gidaris2018unsupervised}, and solving a jigsaw puzzle~\cite{noroozi2016unsupervised}. More recently, contrastive learning methods significantly advanced self-supervised representation learning. The core idea of these approaches is to learn representations invariant to a set of data transformations, \eg, random cropping, random color jittering, and random Gaussian blurring.
A large number of contrastive learning approaches~\cite{wu2018unsupervised,oord2018representation,tian2019contrastive,he2020momentum,chen2020simple,Zheng_2021_ICCV,zheng2021ressl} are based the the instance discrimination framework, where the models learn to maximize the similarity between positive samples while minimize the similarity between the negative samples.
Because instance discrimination relies on a large negative sample size to obtain good performance, some other works proposed to get rid of negative samples by, \eg, incorporating clustering algorithms~\cite{caron2018deep,asano2020self,li2020prototypical,caron2020unsupervised}, simply predicting the representation of one augmented view from the other augmented view of the same image~\cite{grill2020bootstrap,chen2020exploring,huang2021self}, feature decorrelation~\cite{zbontar2021barlow}, or self-distillation~\cite{caron2021emerging}. However, most of the state-of-the-art contrastive learning approaches operate on the global feature only, which risks involving object-irrelevant nuisances and spatial misalignment between different augmented view. The proposed LEWEL framework falls into the category of contrastive learning, addressing the aforementioned drawbacks adaptively.

There have been several researches that dedicated to alleviating these problems. Most of them proposed to maintain the spatial resolution of backbone features and involve the priors of downstream tasks in the pre-training stage. For example, some works performed pixel-level contrastive learning by matching the most similar pixel~\cite{wang2021dense} or the pixels lie in the neighboring region on the original image~\cite{xie2021propagate,pinheiro2020unsupervised}; the works of~\cite{xiao2021region,roh2021spatially} learned to match the representation of the regions that correspond to the same patch in the original image, with the help of RoI Pooling~\cite{he2015spatial}; the work of~\cite{henaff2021efficient} used a set of pre-computed masks to pool corresponding features for matching. In contrast to these methods using pre-defined matching rules, we propose learning where to learn in self-supervised learning with a set of alignment maps predicted on-the-fly during training. Although the way LEWEL predicts the alignment maps is inspired by the seminal learning paradigm of semantic segmentation~\cite{long2015fully} (\ie, per-pixel classification), we do not involve any prior of semantic segmentation during pre-training, \eg, by enforcing pixel-level consistency. Furthermore, owing to the reinterpretation of the global projection for predicting alignment maps, LEWEL couples the learning of the global representation and the aligned representation and improves their performance simultaneously, while the others suffer from degraded classification performance.

%% file: chaps/approach.tex
\section{Learning Where to Learn}
\label{sec:approach}

In this section, we first introduce the generalized spatial aggregation formulation of contrastive learning. Then, we present our methodology, \textbf{Le}arning \textbf{W}her\textbf{e} to \textbf{L}earn (LEWEL), and finally, we discuss the connections between our method and prior works.

\begin{figure*}
    \vspace{-0.1in}
    \centering
    \includegraphics[width=.9\textwidth]{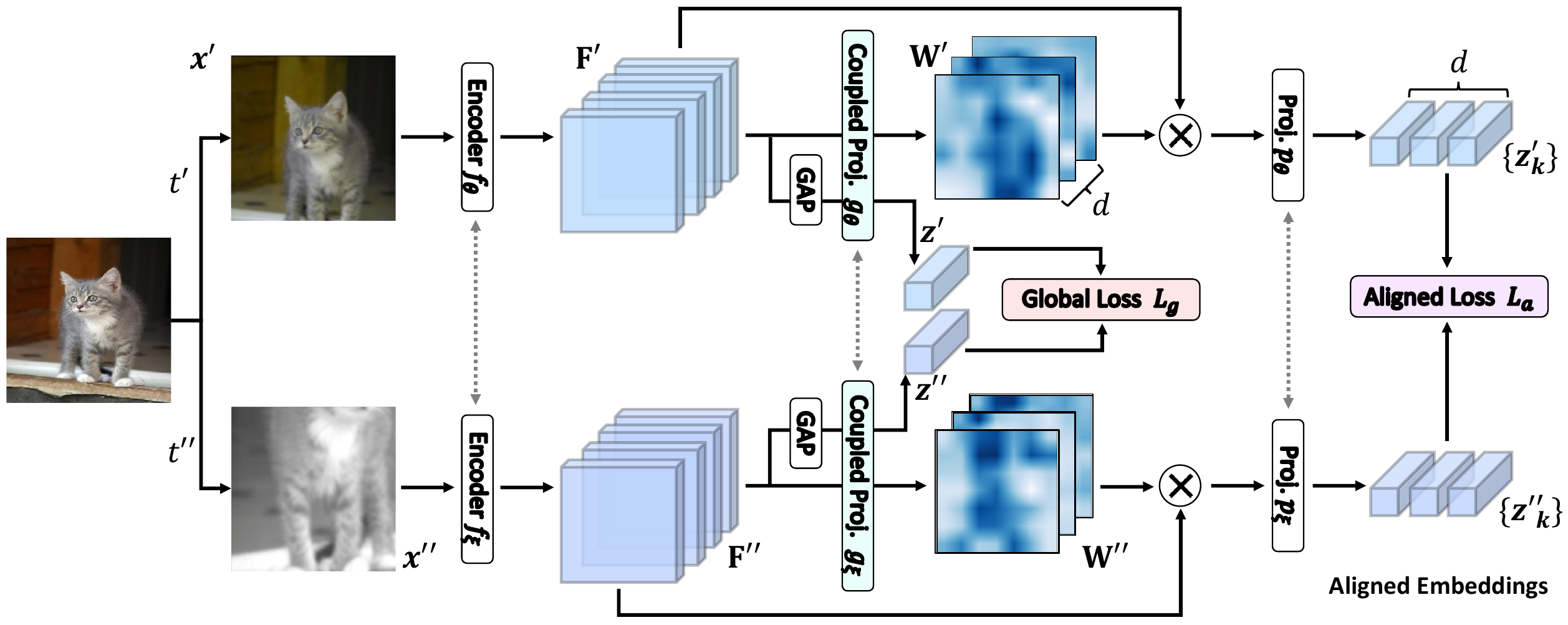}
    \vskip -0.1in
    \caption{\textbf{Overview of the Learning Where to Learn (LEWEL) framework}. Here, $\theta$ denotes the model parameters to be learned and $\xi$ is the exponential-moving average of $\theta$. In LEWEL, each random crop of the original images is independently processed by the encoder $f$, a global average pooling (GAP) layer and a global projector $g$ to produce the global embedding $\z$, upon which the global loss $\mathcal{L}_g$ is applied. In addition, we reinterpret the projection head $g$ as a per-pixel projection to directly predict a set of spatial alignment maps $\mathbf{W}$, \ie coupling the projection head and the alignment map prediction process. Based on the alignment maps, LEWEL adaptively aggregates a spectrum of aligned embeddings $\{\z_k\}$ and minimize the aligned loss $\mathcal{L}_a$ on them. The channel grouping scheme is omitted here for clarity.}
    \label{fig:attn}
    \vspace{-0.2in}
\end{figure*}

\subsection{Generalized spatial aggregation}
\label{sec:appraoch_ours}

\noindent\textbf{Notations.}\quad
Let $\theta$ denote model parameters to be optimized and $\xi$ be the momentum parameters that are updated by exponential moving average (EMA) $\xi\leftarrow \alpha \xi + (1 - \alpha)\theta$. Following this notation, we denote the encoder (e.g., ResNet-50~\cite{he2016deep}) by $f_{\theta}$ and its momentum counterpart by $f_{\xi}$. Besides, we denote the unlabeled training image by $\x$ and a set of data augmentation operations (e.g., random crop, color jitter, etc.) as $\mathcal{T}$, which are used to produce augmented views for self-supervised learning.


\medskip
\noindent\textbf{Global contrastive learning.}\quad
In most of existing contrastive learning frameworks, each training image is independently transformed by two randomly chosen augmentation operations $t', t''\in \mathcal{T}$ to obtain $\x' = t'(\x), \x'' = t''(\x)$. As shown in \cref{fig:reinterpret}a, from the first augmented view $\x'$, the encoder $f_{\theta}$ outputs a backbone feature $\mathbf{F}' = f_{\theta}(\x')$. Then the Global Average Pooling (GAP) operates on the feature $\mathbf{F}'$ to obtain the representation $\y' = \sum_{i=1}^{H} \sum_{j=1}^{W} \frac{1}{H\times W} \mathbf{F}'_{*, i, j} \in \mathbb{R}^{D}$, where $D, H, W$ are the number of channels, height, and width of $\mathbf{F}'$, respectively. 
The representation $\y'$ is further transformed by a projection head $g_{\theta}$ to obtain the embedding $\z' = g_{\theta}(\y') \in \mathbb{R}^{d}$ where $d$ is the dimensionality of $\z'$. Likewise, the backbone feature $\mathbf{F}''$, the representation $\y''$ and the embedding $\z''$ can be produced from the second augmented view $\x''$ by momentum encoder $f_{\xi}$, the GAP operation and the the momentum projection head $g_{\xi}$. Finally, a self-supervised loss $\mathcal{L}_g = \ell(\z', \z'')$ is applied on the two embeddings to drive the training, where the instantiation of $\ell(\cdot, \cdot)$ will be presented subsequently.

\medskip
\noindent\textbf{Learning where to learn.}\quad
The GAP operation is an effective manner of spatial aggregation and introduces the translation invariance. On the one hand, this invariance is favorable for image-level prediction; on the other hand, averaging over all spatial positions induces object-irrelevant nuisances (\eg, background information) and suffers from the loss of spatial information, which is crucial for spatial-sensitive tasks.

In this paper, we propose learning where to learn in self-supervised learning automatically. The formulation of our method is defined as learning an alignment map $\mathbf{W}' \in \mathbb{R}^{H\times W}$ for spatial aggregation:
\begin{align}
\label{eq:aggregation}
    \y' = \mathbf{W}'\bm{\otimes} \mathbf{F}' = \sum_{i=1}^{H} \sum_{j=1}^{W} \mathbf{W}'_{i, j}\mathbf{F}'_{*, i, j},
\end{align}
where $\sum_{i=1}^{H} \sum_{j=1}^{W} \mathbf{W}'_{i, j} = 1, \mathbf{W}'_{i, j} \geq 0$, and $\y' \in \mathbb{R}^{D}$. \cref{eq:aggregation} is a general formulation and allows the model to align the mismatched augmented views, excluding the undesirable nuisances. Furthermore, we can see that the GAP is a special case of \cref{eq:aggregation} where $\mathbf{W}'_{i,j} = \frac{1}{H\times W},~\forall~i,j$, suggesting that our method is able to aggregate both local and global representations in a unified framework by manipulating the alignment map $\mathbf{W}'$.

\subsection{Reinterpreting and coupling projection head}
\noindent\textbf{Reinterpreting and coupling.}\quad
For the global contrastive methods, given an input image $\x'$, the models output a global representation $\y'$ and an embedding $\z' \in \mathbb{R}^d$. This embedding $\z$ consists of $d$ scalars, of which the $k$th element records the activation of image $\x'$ to the $k$th semantic. Here, the concept of ``semantic'' is loosely defined and can represent an object, a pattern, or something else encoded by the model. From this point of view, the projection head $g_{\theta}$ acts like a ``classifier'' atop the global representation $\y'$. With this notion, our methodology is inspired by the learning paradigm of semantic segmentation, \ie, per-pixel classification~\cite{long2015fully}. We reinterpret the global projection head $g$ as a per-pixel projection and apply $g_{\theta}$ to the feature before GAP (\ie, $\mathbf{F}'$). Then, as illustrated in \cref{fig:reinterpret}b, when fed the first augmented view $\x'$, the model outputs $\widetilde{\mathbf{W}}' = g_{\theta}(\mathbf{F}') \in \mathbb{R}^{d\times H\times W}$, containing a set of heat-maps that record the activation of all $H\times W$ positions to all $d$ semantics. This reinterpretation couples the global projection and the alignment map prediction by \textbf{weight sharing} of $g_{\theta}$, allowing the models to learn a better global representation and aligned representation simultaneously.

As in global contrastive learning, the heat-maps $\widetilde{\mathbf{W}}'$ are then $\ell_2$-normalized along the \emph{channel} dimension as $\overline{\mathbf{W}}'_{*,i,j} = \frac{\widetilde{\mathbf{W}}'_{*,i,j}}{ ||\widetilde{\mathbf{W}}'_{*,i,j}||_2},~\forall~i,j$, independently for each position.
Finally, to obtain the alignment maps $\mathbf{W}'$, we normalize each heat-map from $\overline{\mathbf{W}}'$ independently along \emph{spatial} dimensions as $\mathbf{W}'_k = \mathrm{softmax}(\overline{\mathbf{W}}'_k) \in \mathbb{R}^{H\times W}$,
where the softmax function operates on both height and width dimensions, \ie, $\mathbf{W}'_{k,i,j} = \frac{\exp(\overline{\mathbf{W}}'_{k,i,j})}{\sum_{u=1}^{H} \sum_{v=1}^{W} \exp(\overline{\mathbf{W}}'_{k,u,v})},~\forall~i, j$.

\medskip
\noindent\textbf{Channel grouping.}\quad
Instead of using one alignment map to aggregate one aligned representation, we introduce a grouping scheme that divides the channels of $\mathbf{F}'$ uniformly into $h$ equal-size groups, that is $\mathbf{F}' = [\mathbf{F}'^{(1)}, \cdots, \mathbf{F}'^{(h)}]$ where $[\cdot]$ denotes the concatenation operation. 
Given the alignment maps $\{\mathbf{W}'_k\}_{k=1}^{d}$, we can accordingly aggregate a set of aligned representations $\{\y'_k : \y'_k \in \mathbb{R}^{D}\}_{k=1}^{d / h}$, and
{\small{
\begin{align}
\label{eq:aggregate_sem}
    \y'_k = [\mathbf{W}'_{(k-1)\times h+1}\bm{\otimes}\mathbf{F}'^{(1)}, \cdots, \mathbf{W}'_{k\times h}\bm{\otimes}\mathbf{F}'^{(h)}], \forall k,
\end{align}}}
\hspace{-0.05in}where the operation $\bm{\otimes}$ is the spatial aggregation operation defined in \cref{eq:aggregation}. We provide the diagram of the grouping scheme in Appendix~\ref{sec:appendix_grouping} for an intuitive illustration.
This grouping scheme allows us to explicitly control the number of aligned representations while encoding more semantics into each aligned representation. The aligned representations are then projected by a semantic projector $p_{\theta}$ to obtain the aligned embeddings $\{\z'_k: \z'_k = p_{\theta}(\y'_k) \in \mathbb{R}^{c}\}_{k=1}^{d / h}$, where $c$ is the output dimensionality of the projector $p_{\theta}$. Following the same procedure, the aligned embeddings $\{\z''_k\}_{k=1}^{d / h}$ of the second augmented view could be produced according to \cref{eq:aggregate_sem} and the momentum projectors $g_{\xi}$ and $p_{\xi}$. Finally, a self-supervised loss $\mathcal{L}_a = \frac{h}{d}\sum_{k=1}^{d / h} \ell(\z'_k, \z''_k)$ is applied on the aligned embeddings to drive the training.

\subsection{Implementations}
\label{sec:appr_impl}
\noindent\textbf{Loss functions.}\quad
Putting everything together, we formulate the self-supervised learning as minimizing the following objective:
\begin{align}
\label{eq:loss_sum}
\mathcal{L} = (1 - \beta) \mathcal{L}_g + \beta \mathcal{L}_a,
\end{align}
where $\beta$ is the trade-off term between the global loss and aligned loss, which is set to 0.5 by default throughout this paper. The overall pipeline of LEWEL is displayed in \cref{fig:attn} for a more intuitive illustration.

The above formulation is general and is agnostic to the specific choice of the self-supervised loss. 
Here, we present two instantiations of the loss function. The first variant is termed \textbf{LEWEL}{\bm{$_M$}} and is based on the InfoNCE loss~\cite{oord2018representation}:
\begin{align}
\label{eq:contra_formul}
\ell_{\mathrm{InfoNCE}}(\z', \z'') = -\log \frac{e^{\mathrm{sim}(\z', \z'') / \tau}}{e^ {\mathrm{sim}(\z', \z'') / \tau + \sum_{\z^{-}} \mathrm{sim}(\z', \z^{-}) / \tau}},
\end{align}
where $\mathrm{sim}(\cdot, \cdot)$ is the cosine similarity of an input pair, $\tau$ is a temperature term, $\z^-$ is the embedding of a negative sample. In our implementation, the negative samples are stored in a first-in-first-out queue~\cite{he2020momentum} for the global loss $\mathcal{L}_g$, or consists of embeddings of different images in the current mini-batch for $\mathcal{L}_a$ since the quantity is sufficiently large.
Moreover, we present a second variant, termed \textbf{LEWEL}{\bm{$_B$}}, that does not rely on the negative samples. LEWEL$_B$ adopts the normalized Mean Square Error as in BYOL~\cite{grill2020bootstrap}:
\begin{align}
\label{eq:cosine_formul}
\ell_{\mathrm{MSE}}(\z', \z'') = 2 - 2\times \mathrm{sim}(q_{\theta}(\z'), \operatorname{sg}(\z'')),
\end{align}
where $\operatorname{sg}$ stands for the stop-gradient operation. The $q_{\theta}$ is an additional predictor that facilitates the learning of the global representations, while for the aligned representations, a separate predictor $s_{\theta}$ is adopted for the same purpose.

\medskip
\noindent\textbf{Data augmentations.}\quad
LEWEL adopts the same configuration of data augmentations as MoCo~\cite{he2020momentum} and BYOL~\cite{grill2020bootstrap}. In general, we first take two random crops from an input image and resize each of them to 224$\times$224. Then the following random distortions are applied on each crop independently with some probabilities: horizontal flipping, color jittering, converting to grayscale, Gaussian blurring, and (for LEVEL$_B$) solarization. 

\medskip
\noindent\textbf{Architecture.}\quad
We instantiate the encoder $f(\cdot)$ with the ResNet-50~\cite{he2016deep}, which is the most common choice in this literature. The projectors $g(\cdot)$ and $p(\cdot)$ are implemented by the multi-layer perceptrons (MLPs), with one hidden layer followed by a Batch Normalization~\cite{ioffe2015batch} (BN) layer and the ReLU~\cite{nair2010rectified} non-linear activation. The hidden/output dimension of the projectors $g(\cdot)$ and $p(\cdot)$ are set to 2048/128 for LEWEL$_M$ and 4096/256 for LEWEL$_B$, which means that $d$ is 128/256 for LEWEL$_M$/LEWEL$_B$ and $d = c$ by default. Besides, the extra predictors $q(\cdot)$ and $s(\cdot)$ of LEWEL$_B$ are also instantiated by two-layer MLPs, with the same architecture as the projectors.
We follow the settings of MoCo~\cite{he2020momentum} to use the ShuffleBN in the momentum encoder $f_{\xi}$. For LEWEL$_B$, we use the EMA BN statistics in the momentum encoder by default for faster training, following~\cite{cai2021exponential}. With 400 training epochs, we use the SyncBN for better performance, following BYOL~\cite{grill2020bootstrap}.


\medskip
\noindent\textbf{Optimization.}\quad
We use the Stochastic Gradient Descent (SGD) with the momentum of 0.9 to minimize our objective functions. For LEWEL$_M$, we use the batch size of 256, initial learning rate of 0.06, weight decay of $10^{-4}$, temperature term $\tau$ of 0.2, and fix the momentum $\alpha$ of encoder to 0.999. For LEWEL$_B$, we use the batch size of 512, initial learning rate of 1.8, weight decay of $10^{-6}$, and set the initial momentum $\alpha$ of the encoder to 0.98, which is increased to 1 according to the cosine schedule~\cite{grill2020bootstrap}. By default, we perform self-supervised pre-training on the ImageNet-1K~\cite{russakovsky2015imagenet} (IN-1K) datasets using a single machine with eight GPUs.
The models are trained for 100/200/400 epochs using the cosine annealing schedule~\cite{loshchilov2016sgdr} and Automatic Mixed-Precision training on PyTorch~\cite{paszke2019pytorch}. For LEWEL$_B$, the weight decay of bias and BN layers are set to 0.
In our ablation studies, we pre-train the models on the ImageNet-100~\cite{tian2019contrastive} (IN-100) dataset for fast iteration. In such a case, we simply double the initial learning rate of LEWEL and train the models for 240 epochs following~\cite{tian2019contrastive} and keep the rest unchanged.

\subsection{Discussions}
\label{sec:appr_discs}
\noindent\textbf{From the spatial alignment perspective.}\quad
The formulation described in \cref{eq:aggregation} generalizes to most of prior SSL methods, including the GAP-based approaches and the pixel-based or patch-based approaches. For example, the pixel-based approaches, whose objective is to learn pixel-wise correspondence, essentially find a set of one-hot alignment maps that each activates on the single corresponding position only; the patch-based methods take a set of pre-defined alignment maps that activates the patch regions only for spatial alignment. In sharp contrast to these methods, we propose automatically learning where to learn on the models' own, according to the general formulation \cref{eq:aggregation}. Moreover, since no prior of a specific downstream task is involved in the pre-training stage, LEWEL is able to perform well on both image-level and dense predictions, rather than trading-off the performance of one task to the others'.

\medskip
\noindent\textbf{From the embedding perspective.}\quad
For the global contrastive methods, given an input image $\x$, the models output an global embedding $\z \in \mathbb{R}^d$. From the embedding perspective, this embedding $\z$ consists of $d$ scalars, of which the $k$th element records the activation of the input images $\x$ to the $k$th semantic. By contrast, guided by the semantic-aware alignment maps, LEWEL (with the number of groups $h=1$) encodes $d$ aligned embeddings for all semantics accordingly, each of which has a dimensionality of $c$. In other words, under our framework, more expressive powers are assigned to each semantic by the aligned embeddings, so that the learning on the aligned embeddings implicitly benefits the learning of global embedding $\z$. We will discuss the influence of $c$ in \cref{sec:exp_abla}.

%% file: chaps/exp.tex
\section{Experiments}
\label{sec:exp}

\begin{table}
  \centering
  \small
  \caption{\textbf{Comparison on IN-1K linear classification} with the ResNet-50 models pre-trained on the IN-1K dataset. $\dag$: results cited from~\cite{chen2020exploring}. $*$: our reproduction.
  }
  \label{tab:exp_in1k_cls}
  \vspace{-0.1in}
  \setlength{\tabcolsep}{1.25mm}{
  \begin{tabular}{@{}lccccc@{}}
    \toprule
    \multirow{2}{*}{Method} & \multicolumn{2}{c}{100 Epochs} & \multicolumn{2}{c}{200 Epochs} & 400 Epochs \\
     \cmidrule(l{3pt}r{3pt}){2-3} \cmidrule(l{3pt}r{3pt}){4-5} \cmidrule(l{3pt}r{3pt}){6-6}
     & Acc@1 & Acc@5 & Acc@1 & Acc@5 & Acc@1 \\
    \midrule
    InstDisc~\cite{wu2018unsupervised} & - & - & 56.5 & - & -\\
    PCL~\cite{li2020prototypical} & - & - & 67.6 & - & -\\
    SimCLR~\cite{chen2020simple} &  64.6 & - & 66.6 & - & -\\
    SimCLR~\cite{chen2020simple}$^\dag$ & 66.5 & - & 68.3 & - & 70.4 \\
    BYOL~\cite{grill2020bootstrap}$^\dag$ & 66.5 & - & 70.6 & - & 73.2 \\
    SwAV~\cite{caron2020unsupervised}$^\dag$ & 66.5 & - & 69.1 & - & 70.7 \\
    SimSiam~\cite{chen2020exploring}$^\dag$ & 68.1 & - & 70.0 & -&  70.8 \\
    MoCov2~\cite{he2020momentum}$^*$ & 64.5 & 86.1 & 67.5 & 88.1 & - \\
    BYOL~\cite{grill2020bootstrap}$^*$ & 70.6 & 89.9 & 71.9 & 90.4 & - \\
    \rowcolor{Gray}
    LEWEL$_{M}$ & 66.1 & 87.2 & 68.4 & 88.6 & - \\
    \rowcolor{Gray}
    LEWEL$_{B}$ & \textbf{71.9} & \textbf{90.5} & \textbf{72.8} & \textbf{91.0} & \textbf{73.8} \\
    \bottomrule
  \end{tabular}
  }
  \vskip -0.2in
\end{table}

\subsection{Linear evaluation}
\label{sec_exp_lin_cls}
\noindent\textbf{Experimental setup.}\quad
Following prior works~\cite{chen2020exploring,grill2020bootstrap}, we remove the projectors and predictors in LEWEL and train a linear classifier atop the fixed backbone $f_{\theta}$ to evaluate the learned representations. For LEWEL$_M$, we train the linear classifier for 90 epochs with batch size 4,096, initial learning rate 3.2, weight decay 0, LARS optimizer~\cite{you2017large} and the cosine annealing schedule~\cite{loshchilov2016sgdr}, following~\cite{chen2020exploring}. For LEWEL$_B$, we train the classifier for 50 epochs with batch size 256, weight decay 0, SGD optimizer with momentum, learning rate 0.4 that is decayed by a factor of 10 at the 30th and 40th epoch, following~\cite{cai2021exponential}.

\medskip
\noindent\textbf{Results.}\quad
The top1 and top5 validation accuracy on IN-1K are reported in \cref{tab:exp_in1k_cls}, which includes both cited results and our reproduced results for fair comparisons. Overall, the proposed LEWEL outperforms the state-of-the-art methods under all settings by a substantial margin: with 100 training epochs, LEWEL$_M$/LEWEL$_B$ improve the top1 validation accuracy of their baseline methods MoCo~\cite{he2020momentum}/BYOL~\cite{grill2020bootstrap} by 1.6\%/1.3\% points; with 400 training epochs, LEWEL$_B$ outperforms BYOL by 0.6\% points.

\begin{table}
  \centering
  \small
  \caption{\textbf{Comparison on IN-1K semi-supervised classification} with the ResNet-50 models pre-trained on the IN-1K dataset. $*$: our reproductions.}
  \label{tab:exp_in1k_semi_cls}
  \vskip -0.1in
  \setlength{\tabcolsep}{1.7mm}{
  \begin{tabular}{@{}lccccc@{}}
    \toprule
    \multirow{2}{*}{Method} & \multirow{2}{*}{Epochs} & \multicolumn{2}{c}{1\% Labels} & \multicolumn{2}{c}{10\% Labels} \\
     \cmidrule(l{3pt}r{3pt}){3-4} \cmidrule(l{3pt}r{3pt}){5-6}
     & & Acc@1 & Acc@5 & Acc@1 & Acc@5 \\
    \midrule
    PCL~\cite{li2020prototypical} & 200 & - & 75.3 & - & 86.5 \\
    MoCov2~\cite{he2020momentum}$^*$ & 200 & 43.8 & 72.3 & 61.9 & 84.6 \\
    BYOL~\cite{grill2020bootstrap}$^*$ & 200 & 54.8 & 78.8 & 68.0 & 88.5 \\
    \rowcolor{Gray}
    LEWEL$_{M}$ & 200 & 45.1 & 71.1 & 62.5 & 84.9 \\
    \rowcolor{Gray}
    LEWEL$_{B}$ & 200 & \textbf{56.1} & \textbf{79.9} & \textbf{68.7} & \textbf{88.9} \\
    \midrule
    SimCLR~\cite{chen2020simple} & 1000 & 48.3 & 75.5 & 65.6 & 87.8\\
    SwAV~\cite{caron2020unsupervised} & 800 & 53.9 & 78.5 & 70.2 & 89.9 \\
    BYOL~\cite{grill2020bootstrap} & 800 & 53.2 & 78.4 & 68.8 & 89.0 \\
    BarlowTw.~\cite{zbontar2021barlow} & 1000 & 55.0  & 79.2 & 69.7 & 89.3 \\
    \rowcolor{Gray}
    LEWEL$_{B}$ & 400 & \textbf{59.8} & \textbf{83.2} & \textbf{70.4} & \textbf{90.1} \\
    \bottomrule
  \end{tabular}
  }
  \vskip -0.2in
\end{table}

\subsection{Semi-supervised classification}
\noindent\textbf{Experimental setup.}\quad
We further evaluate the fine-tuning performance of the self-supervised pre-trained ResNet-50 on subsets of the IN-1K data. For a fair comparison, we use the 1\% and 10\% subsets that are randomly selected by Chen et al.~\cite{chen2020simple}. We fine-tune the models on these two subsets for 50 epochs with classifier learning rate 1.0 (0.1), backbone learning rate 0.0001 (0.01) for the 1\% (10\%) subset, which are decayed by a factor of 10 at the 30th and 40th epoch.

\medskip
\noindent\textbf{Results.}\quad
The top1 and top5 semi-supervised classification accuracy on the IN-1K validation set are reported in \cref{tab:exp_in1k_semi_cls}. Using the same pre-training epochs, LEWEL outperforms the other methods by a noticeable margin: in particular, when only 1\% of labels are available, LEWEL$_B$ achieves 56.1\%/79.9\% top1/top5 accuracy, improving the other methods by up to 1.3\%/1.1\% points under the setting of 200 pre-training epochs. Furthermore, we find that LEWEL$_B$ outperforms state-of-the-art SSL approaches with $2 \times$ or more pre-training epochs, \eg, LEWEL${_B}$ with 400 epochs perform clearly better than BYOL with 800 epochs. The results of semi-supervised classification and linear classification suggest that LEWEL learns better representations for image-level prediction.

\subsection{Transfer learning to other tasks}
\label{sec_exp_transfer}
To evaluate the transfer learning performance of the pre-trained model to other tasks, we use two standard benchmarks: Pascal VOC~\cite{everingham15} and MS-COCO~\cite{lin2014microsoft}.

\begin{table}
  \centering
  \small
  \caption{\textbf{Transfer learning to Pascal-VOC Object Detection and Semantic Segmentation} with models pre-trained on IN-1K datasets. All entries are based on the Faster R-CNN~\cite{ren2015faster} architecture with the ResNet-50 C4 backbone~\cite{wu2019detectron2}. $\dag$: results cited from~\cite{chen2020exploring}. $*$: our reproductions.}
  \label{tab:exp_voc}
  \vskip -0.1in
  \setlength{\tabcolsep}{2mm}{
  \begin{tabular}{@{}lccccc@{}}
    \toprule
    \multirow{2}{*}{Method} & \multirow{2}{*}{Epochs} & \multicolumn{3}{c}{VOC 07+12 Det.} & 12 Seg. \\
     \cmidrule(l{3pt}r{3pt}){3-5} \cmidrule(l{3pt}r{3pt}){6-6}
     & & AP & AP$_{50}$ & AP$_{75}$ & mIoU \\
    \midrule
    Supervised$^\dag$ & 90 & 53.5 & 81.3 & 58.8 & 67.7 \\
    \midrule
    MoCov2~\cite{he2020momentum}$^*$ & 100 & 56.1 & 81.5 & 62.4 & 66.3 \\
    BYOL~\cite{grill2020bootstrap}$^*$ & 100 & 55.5 & 81.9 & 61.2 & 66.9 \\
    \rowcolor{Gray}
    LEWEL$_{M}$ & 100 & \textbf{56.5} & \textbf{82.1} & \textbf{63.0} & 66.8 \\
    \rowcolor{Gray}
    LEWEL$_{B}$ & 100 & 56.1 & \textbf{82.1} & 62.3 & \textbf{67.6} \\
    \midrule
    SimCLR~\cite{chen2020simple}$^\dag$ & 200 & 55.5 & 81.8 & 61.4 & - \\
    SwAV~\cite{caron2018deep}$^\dag$ & 200 & 55.4 & 81.5 & 61.4 & - \\
    BYOL~\cite{grill2020bootstrap}$^\dag$ & 200 & 55.3 & 81.4 & 61.1 & - \\
    SimSiam~\cite{chen2020exploring}$^\dag$ & 200 & 56.4 & 82.0 & 62.8 & - \\
    MoCov2~\cite{he2020momentum}$^*$ & 200 & 57.0 & 82.2 & 63.4 & 66.7 \\
    BYOL~\cite{grill2020bootstrap}$^*$ & 200 & 55.8 & 81.6 & 61.6 & 67.2 \\
    \rowcolor{Gray}
    LEWEL$_{M}$ & 200 & \textbf{57.3} & 82.3 & 63.6 & 67.2 \\
    \rowcolor{Gray}
    LEWEL$_{B}$ & 200 & 56.5 & \textbf{82.6} & \textbf{63.7} & \textbf{67.8} \\
    \bottomrule
  \end{tabular}
  }
  \vskip -0.2in
\end{table}

\medskip
\noindent\textbf{VOC Object Detection and Semantic Segmentation.}\quad
For object detection, we use the pre-trained model to initialize the ResNet-50-C4 backbone of the Faster-RCNN~\cite{ren2015faster} model. The models are trained on the $\operatorname{trainval07+12}$ split ($\sim$16.5k images) and evaluated on the $\operatorname{test12}$ split ($\sim$5k images), using the opensource codebase detectron2~\cite{wu2019detectron2}. We follow the standard schedule in~\cite{wu2019detectron2}, \ie, 24k iterations with a batch size of 16, decaying the learning rate at $3/4$ and $11/12$ of the total steps, and using SyncBN. For the semantic segmentation, we use the dilated FCN~\cite{long2015fully} model with output stride of 8, which is trained on Pascal VOC 2012 $\operatorname{train+aug}$ split ($\sim$10.6k images) and evaluted on the $\operatorname{val}$ split ($\sim$1.5k images) using the mmsegmentation~\cite{mmseg2020} codebase. We train the models for 20k iteration with a batch size of 16, SyncBN and the ``poly'' learning rate schedule~\cite{chen2017deeplab}. For all models, we search for the best fine-tuning learning rate and report the corresponding results.

The experimental results of object detection (measured by Average Precision (AP), AP$_{50}$, and AP$_{75}$) and semantic segmentation (measured by mean-Intersection-of-Union (mIoU)) are summarized in \cref{tab:exp_voc}. Though the prior work~\cite{chen2020improved} reported that the linear accuracy is not necessarily related to the performance on downstream task, we observe that LEWEL successfully achieves non-trivial improvements on both object detection and semantic segmentation. In fact, LEWEL outperforms all compared methods on almost all entries, using 100-/200-epoch training budget.

\begin{table}
  \vskip -0.075in
  \centering
  \small
  \caption{\textbf{Transfer learning to MS-COCO Object Detection and Instance Segmentation} with models pre-trained for 200 epochs on IN-1K dataset. All entries are based on the Mask R-CNN~\cite{he2017mask} architecture. $\dag$: results from~\cite{chen2020exploring}. $*$: our reproduction.}
  \label{tab:exp_coco}
  \vskip -0.1in
  \setlength{\tabcolsep}{1.25mm}{
  \begin{tabular}{@{}lcccccc@{}}
    \toprule
    \multirow{2}{*}{Method} & \multicolumn{3}{c}{Object Det.} & \multicolumn{3}{c}{Instance Seg.} \\
     \cmidrule(l{3pt}r{3pt}){2-4} \cmidrule(l{3pt}r{3pt}){5-7}
     & AP & AP$_{50}$ & AP$_{75}$ & AP & AP$_{50}$ & AP$_{75}$ \\
    \midrule\midrule
    \multicolumn{7}{l}{\hspace{-0.05in}\tt{ResNet50-C4:}} \\
    Supervised$^\dag$ & 38.2 & 58.2 & 41.2 & 33.3 & 54.7 & 35.2 \\
    SimCLR~\cite{chen2020simple}$^\dag$ & 37.9 & 57.7 & 40.9 & 33.3 & 54.6 & 35.3 \\
    SwAV~\cite{caron2018deep}$^\dag$ & 37.6 & 57.6 &  40.3 & 33.1 & 54.2 & 35.1 \\
    BYOL~\cite{grill2020bootstrap}$^\dag$ & 37.9 & 57.8 & 40.9 & 33.2 & 54.3 & 35.0 \\
    SimSiam~\cite{chen2020exploring}$^\dag$ & 37.9 & 57.5 & 40.9 & 33.2 & 54.2 & 35.2 \\
    MoCov2~\cite{he2020momentum}$^*$ & 38.8 & 58.0 & \textbf{42.0} & 34.0 & 55.2 & \textbf{36.3} \\
    BYOL~\cite{grill2020bootstrap}$^*$ & 38.1 & 58.4 & 40.9 & 33.3 & 55.0 & 35.3 \\
    \rowcolor{Gray}
    LEWEL$_{M}$ & \textbf{38.9} & 58.6 & \textbf{42.0} & \textbf{34.1} & 55.3 & \textbf{36.3} \\
    \rowcolor{Gray}
    LEWEL$_{B}$ & 38.5 & \textbf{58.9} & 41.2 & 33.7 & \textbf{55.5} & 35.5 \\
    \midrule\midrule
    \multicolumn{7}{l}{\hspace{-0.05in}\tt{ResNet50-FPN:}} \\
    DenseCL~\cite{wang2021dense} & 40.3 & 59.9 & 44.3 & 36.4 & 57.0 & 39.2 \\
    ReSim~\cite{xiao2021region} & 39.8 & 60.2 & 43.5 & 36.0 & 57.1 & 38.6 \\
    \rowcolor{Gray}
    LEWEL$_{M}$ & 40.0 & 59.8 & 43.7 & 36.1 & 57.0 & 38.7 \\
    \rowcolor{Gray}
    LEWEL$_{B}$ & \textbf{41.3} & \textbf{61.2} & \textbf{45.4} & \textbf{37.4} & \textbf{58.3} &\textbf{40.3} \\
    PixelPro~\cite{xie2021propagate} (400 ep) & 41.4 & 61.6 & 45.4 & 37.4 & - & - \\
    \rowcolor{Gray}
    LEWEL$_B$ (400 ep) & \textbf{41.9} & \textbf{62.4} & \textbf{46.0} & \textbf{37.9} & \textbf{59.3} & \textbf{40.7} \\
    \midrule\bottomrule
  \end{tabular}
  }
  \vspace{-0.2in}
\end{table}

\medskip
\noindent\textbf{COCO Object Detection and Instance Segmentation.}\quad
We adopt the Mask R-CNN~\cite{he2017mask} architecture with the ResNet50-C4~\cite{wu2019detectron2} (following~\cite{he2020momentum,chen2020exploring}) or ResNet50-FPN~\cite{lin2017feature} (following~\cite{wang2021dense,xie2021propagate,xiao2021region}) backbone, which is pre-trained for 200/400 epochs on ImageNet-1K dataset. All models are fine-tuned on the COCO 2017 $\operatorname{train}$ split ($\sim$118k images) and finally evaluated on the $\operatorname{val}$ split ($\sim$5k images). We use a batch size of 16 and adopt the $1\times$ schedule in the detetron2~\cite{wu2019detectron2}, which uses 90k training iterations in total and decays the learning rate at the 60k-th and 80k-th iteration by a factor of 10. We search the fine-tuning learning rate for LEWEL and the reproduced methods.

The standard COCO metrics, including AP, AP$_{50}$ and AP$_{75}$ for both object detection and instance segmentation, of all methods are reported in \cref{tab:exp_coco}.
We can see that LEWEL achieves the best performance in terms of all metrics. Concretely, LEWEL$_M$/LEWEL$_B$ consistently improve the strong baselines MoCo/BYOL on all entries by up to 0.5\% points. And the gains become even larger when compared with other state-of-the art methods without spatial alignment.
Furthermore, compared with the hand-crafted spatial alignment methods~\cite{wang2021dense,xie2021propagate,xiao2021region}, our experiments in Appendix~\ref{sec:appendix_spt_ali} show that LEWEL 1) performs on par with or even better than them on the dense prediction tasks under both $1\times$ and $2\times$ finetuning schedules; 2) significantly outperforms them on classification.
These experiments, combined with those on Pascal-VOC benchmark, clearly demonstrate that LEWEL is able to improve the dense prediction performance of self-supervised learning.

\subsection{Ablations}
\label{sec:exp_abla}
\noindent\textbf{Experimental setups.}\quad
We pre-train all the models on IN-100 dataset, a subset of the IN-1K selected by \cite{tian2019contrastive}. The models are evaluated with the linear evaluation and on the semantic segmentation, as described in Secs.~\ref{sec_exp_lin_cls} and \ref{sec_exp_transfer}. By default, we use LEWEL$_M$ because it is faster to train.

\begin{table}
  \vskip -0.075in
  \centering
  \small
  \caption{\textbf{The influence of each component} on IN-100 linear classification and transfer learning to VOC 12 semantic segmentation.
  }
  \label{tab:exp_abla_cpn}
  \vskip -0.1in
  \setlength{\tabcolsep}{1.mm}{
  \begin{tabular}{@{}ccccc@{}}
    \toprule
    Global & Align. & Coupled Head & IN-100 Acc. & VOC Seg. mIoU \\
    \midrule
    $\checkmark$ & $\times$      & $\times$ & 79.5 & 61.6 \\
    $\times$     & $\checkmark$  & $\times$ & 80.0 & 62.6 \\
    $\checkmark$ & $\checkmark$  & $\times$ & 81.0 & 62.7 \\
    \rowcolor{Gray}
    $\checkmark$ & $\checkmark$  & $\checkmark$ & \textbf{82.1} & \textbf{63.4} \\
    \bottomrule
  \end{tabular}
  }
  \vskip -0.1in
\end{table}

\begin{figure*}[t]
    \vskip -0.1in
    \centering
    \begin{subfigure}{.48\textwidth}
        \centering
        \includegraphics[width=.49\textwidth]{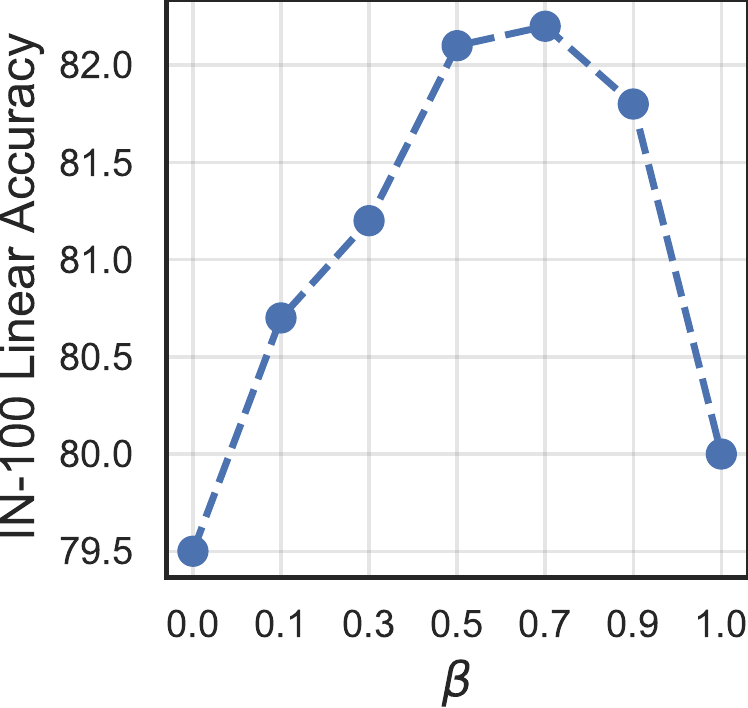}
        \includegraphics[width=.49\textwidth]{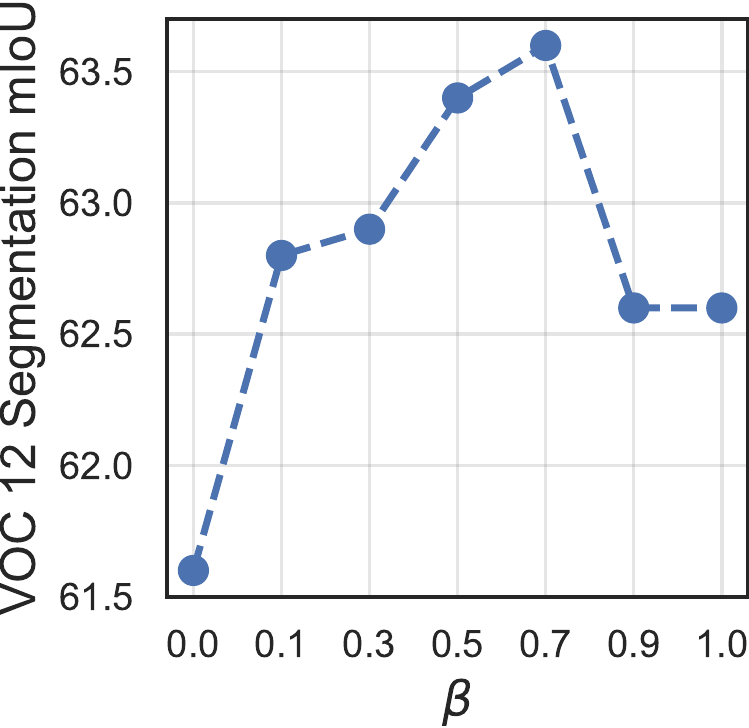}
        \vskip -0.025in
        \caption{The influence of trade-off term $\beta$.}
        \label{fig:exp_loss_weight}
    \end{subfigure}
   \quad
    \begin{subfigure}{.48\textwidth}
        \centering
        \includegraphics[width=.50\textwidth]{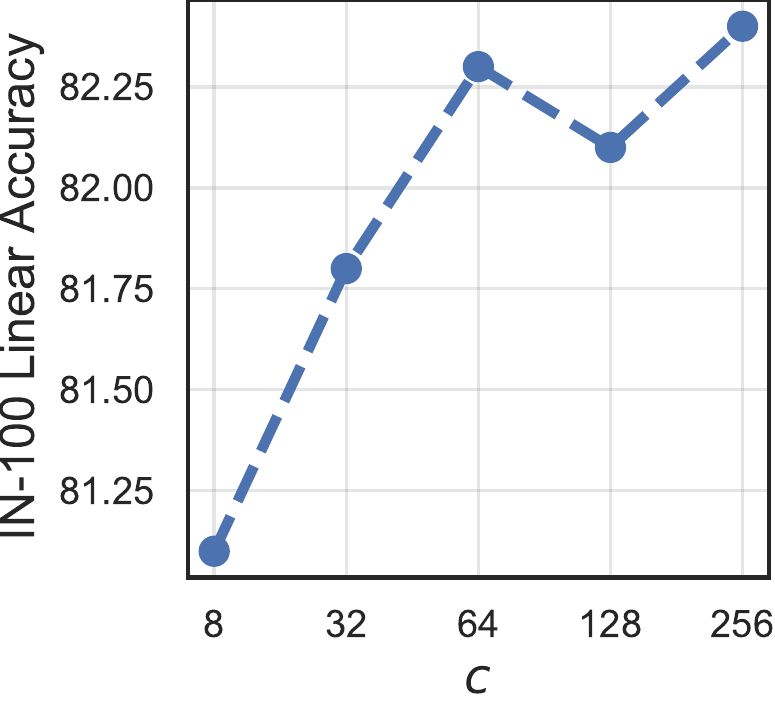}
        \includegraphics[width=.49\textwidth]{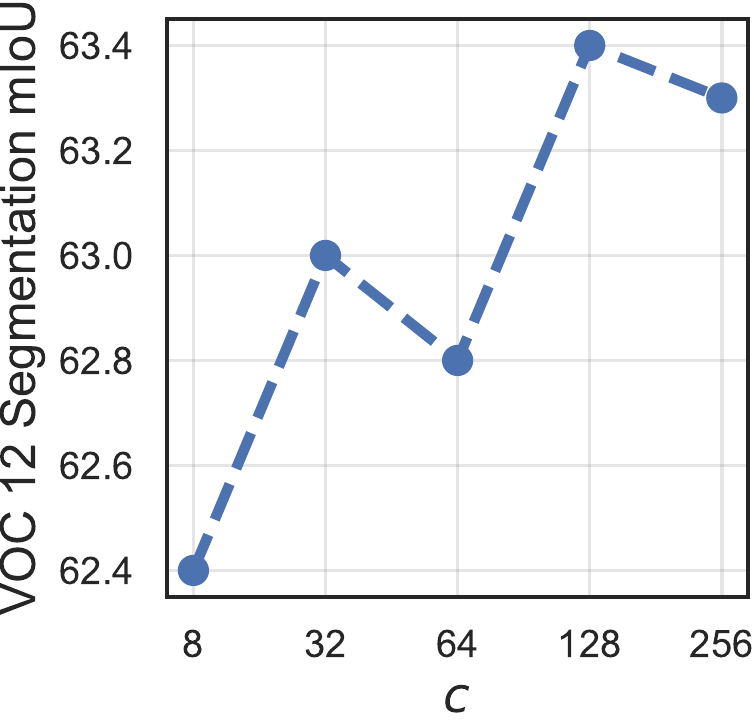}
        \vskip -0.025in
        \caption{The influence of the dimensionality $c$ of the aligned embeddings.}
        \label{fig:exp_abla_aligned_dim}
    \end{subfigure}
    \vspace{-0.125in}
    \caption{\textbf{Ablation studies} of (a) the loss weight $\beta$, and (b) the dimensionality $c$ of the aligned embeddings on IN-100 and VOC 12.
    }
    \label{fig:exp_abla_all}
    \vspace{-0.2in}
\end{figure*}

\medskip
\noindent\textbf{Influence of each component.}\quad
In \cref{tab:exp_abla_cpn}, we investigate the contributions of the introduced aligned loss and weight-sharing scheme of the projector $g$ (\ie the coupled head design) to our approach. We can see that, using only the global loss (\ie, the MoCov2 baseline) or aligned loss, the results on the two evaluated tasks are indeed inferior to the variant using both of these two losses. Moreover, the incorporation of the weight-sharing scheme yields the further gains, improving the vanilla baseline by 2.6\%/1.8\% points in terms of linear classification accuracy/segmentation mIoU. The results verifies the effectiveness of the aligned loss and the coupled head / weight-sharing scheme of LEWEL.

\begin{table}
  \centering
  \small
  \caption{\textbf{Comparison with MoCo with large projection heads} on IN-100 linear classification and VOC semantic segmentation.
  }
  \label{tab:exp_abla_proj_heads_dim}
  \vskip -0.1in
  \setlength{\tabcolsep}{1.0mm}{
  \begin{tabular}{@{}lcccc@{}}
    \toprule
    Method & Extra $p_{\theta}$ & $d$ & IN-100 Acc. & VOC mIoU \\
    \midrule
    \multirow{4}{*}{MoCov2~\cite{he2020momentum}} & $\times$ & 128 & 79.5 & 61.6\\
    & $\times$ & 256 & 79.8 & 60.6 \\
    & $\times$ & 512 & 80.2 & 61.5 \\
    & $\checkmark$ & 128 & 79.9 & 62.2 \\
    \midrule
    LEWEL$_M$ & $\times$ & 128 & 81.8 & 63.0 \\
    LEWEL$_M$ w/ rand. $\mathbf{W}$ & $\checkmark$ & 128 & 79.8 & 61.9 \\
    \rowcolor{Gray} 
    LEWEL$_M$ & $\checkmark$ & 128 & \textbf{82.1} & \textbf{63.4} \\
    \bottomrule
  \end{tabular}
  }
  \vskip -0.2in
\end{table}

\medskip
\noindent\textbf{Comparison to MoCov2 with larger/extra projection heads.}\quad
Since the LEWEL uses a separate projection head for the aligned representation, it is important to clearly identify the contribution of the extra parameters to its performance improvement. In \cref{tab:exp_abla_proj_heads_dim}, we compare the LEWEL$_M$ with MoCov2~\cite{he2020momentum} with larger or extra projection heads. We do not observe significant improvements with neither the larger nor extra projection head for MoCo. Moreover, we notice that a similar study from~\cite{zbontar2021barlow} also suggested that neither deeper nor wider projection head could improve the performance of BYOL. In contrast, switching from MoCo to LEWEL significantly improves the performance on both classification and segmentation, \ie, by 2.6\%/1.8\% points, respectively. The experiments indicate that the improvement of LEWEL mostly comes from the algorithm itself, not from the slightly increased parameters.

\medskip
\noindent\textbf{Influence of the loss weight.}\quad
We also conducted experiments to study to what extent the loss weight $\beta$ affects the performance of LEWEL, though we fix $\beta=0.5$ by default for simplicity. The experimental results are shown in \cref{fig:exp_loss_weight}. We observe that, when $\beta$ is too small (\eg, $\leq 0.3$), the models perform relatively worse than those with a large $\beta$. When $\beta$ is set to a larger value, the models actually produce similar results, except for the case of $\beta = 1$. This observation suggests that our reinterpretation of the global projection head and the weight-sharing scheme couple the learning on the global loss and the aligned loss closely, so that the learning of the aligned representations benefits the learning of the global representation.

\medskip
\noindent\textbf{Influence of the dimensionality of the aligned embeddings.}\quad
As discussed in \cref{sec:appr_discs}, each of the aligned embedding is actually an extra representation for a corresponding semantic encoded by the global embedding.
In \cref{fig:exp_abla_aligned_dim}, we compare the performance of LEWEL$_M$ with different dimensionality $c$ of the aligned embeddings while keep the output dimensionality of the projection head $g(\cdot)$ fixed. From the result, we can see that the performance initially increases w.r.t. the dimensionality $c$ and then stagnates around the point that $c=64$. The results  suggest that the extra expressive power of aligned embedding indeed helps the learning of global representation. However, when the dimensionality is too large, the extra information might be redundant so that the performance will not further increase.

\medskip
\noindent\textbf{Influence of the number of aligned embeddings.}\quad
The grouping scheme allows the models to encode richer semantics into each aligned representation while reduces the number of aligned representations. In \cref{tab:exp_abla_heads}, we compare the performance of the model variants with different numbers of the aligned embeddings, which are determined by the output dimensionality $d$ of the coupled head $g(\cdot)$ and the number of groups $h$. Here, we use the LEWEL$_B$ for this study in order to exclude the influence of the negative sample size in LEWEL$_M$. From the table, we have an intriguing observation when fixing $d=256$: while the linear accuracy increases as $h$ becomes larger and then slight drops, the performance in semantics segmentation exhibits a completely opposite trend that favors a smaller $h$. This observation may suggest that the aligned representations tend to focus (a) on local regions when the number of aligned embeddings is too large, and (b) on the global content when the number is small. We also find that when $h=4$, LEWEL$_B$ performs consistently well regardless the value of $d$. Thus we simply use $d=4$ for LEWEL$_B$ and adopt $d=256$ for direct comparison with the baseline BYOL.

\begin{table}
  \centering
  \small
  \caption{\textbf{The influence of the number of aligned embeddings} on IN-100 linear classification and VOC 12 semantic segmentation.
  }
  \label{tab:exp_abla_heads}
  \vskip -0.1in
  \setlength{\tabcolsep}{2mm}{
  \begin{tabular}{@{}lcccc@{}}
    \toprule
    Method & $d$ & $h$ & IN-100 Acc.  & VOC 12 Seg. mIoU \\
    \midrule 
    BYOL & 256 & N/A & 81.4 & 59.7 \\
    \midrule
    \multirow{8}{*}{LEWEL$_B$} & 512 & 4 & 84.6 & 62.6 \\
    \cmidrule{2-5}
     & 256 & 1 & 81.0 & 63.2 \\
     & 256 & 2 & 82.6 & 63.3 \\
     & \cellcolor{Gray}256 & \cellcolor{Gray}4 & \cellcolor{Gray}\textbf{83.3} & \cellcolor{Gray}\textbf{63.4} \\
     & 256 & 8 & 83.1 & 62.1 \\
     & 256 & 16 & 82.9 & 61.4 \\
     \cmidrule{2-5}
     & 128 & 4 & 83.8 & 60.9 \\
     \cmidrule{2-5}
     & 64 & 4 & 84.8 & 61.8 \\
    \bottomrule
  \end{tabular}
  }
  \vspace{-0.2in}
\end{table}

\medskip
\noindent\textbf{Effectiveness of spatial alignments.}\quad
We report the performance of LEWEL with a random alignment map $\mathbf{W}$ in the second last row of~\cref{tab:exp_abla_proj_heads_dim}, which is significantly lower than the default one, highlighting the importance of adaptive alignment in LEWEL. Moreover, we visualize the alignment map in Appendix~\ref{sec:appendix_vis} to show that LEWEL can automatically find semantically consistent alignments.

%% file: chaps/conclusion.tex
\section{Conclusion}
\label{sec:conclusion}
In this work, we present a new approach, Learning Where to Learn (LEWEL), for self-supervised learning (SSL), which is in sharp contrast to existing SSL methods that learn on a fixed (global or local) region. We reinterpret the global projection head in SSL as per-pixel projection, predicting a set of alignment maps to adaptively aggregate spatial information for SSL. As a result of our adaptive alignment and reinterpretation scheme, we observe significantly improvements of LEWEL over the state-of-the-art SSL methods on various tasks, including linear/semi-supervised classification, object detection, and instance/semantic segmentation.

%% file: chaps/appendix.tex
\begin{table*}[t]
  \centering
  \small
    \caption{\textbf{Comparison with hand-crafted spatial alignment methods}. The experiments on MS-COCO is based on Mask R-CNN architecture with the ResNet-50 FPN backbone and the $1\times$/$2\times$ schedule~\cite{wu2019detectron2}, following~\cite{wang2021dense,xie2021propagate,xiao2021region}.}
  \label{tab:exp_spt_alm}
  \setlength{\tabcolsep}{1.5mm}{
  \begin{tabular}{@{}l|c|cc|cccccc|cccccc@{}}
    \toprule
    \multirow{2}{*}{Method} & \multirow{2}{*}{Epoch} & \multicolumn{2}{c|}{IN-1K} & \multicolumn{6}{c|}{MS-COCO ($1\times$ Schedule)} & \multicolumn{6}{c}{MS-COCO ($2\times$ Schedule)} \\
    \cline{3-4} \cline{5-10}  \cline{11-16}
     & & Acc@1 & Acc@5 & AP$^{\mathrm{b}}$ & AP$^{\mathrm{b}}_{50}$ & AP$^{\mathrm{b}}_{75}$ & AP$^{\mathrm{m}}$ & AP$^{\mathrm{m}}_{50}$ & AP$^{\mathrm{m}}_{75}$ & AP$^{\mathrm{b}}$ & AP$^{\mathrm{b}}_{50}$ & AP$^{\mathrm{b}}_{75}$ & AP$^{\mathrm{m}}$ & AP$^{\mathrm{m}}_{50}$ & AP$^{\mathrm{m}}_{75}$ \\
    \midrule
    DenseCL~\cite{wang2021dense} & 200 & 63.6 & 85.8 & 40.3 & 59.9 & 44.3 & 36.4 & 57.0 & 39.2 & 41.2 & 61.9 & 45.1 & 37.3 & 58.9 & 40.1 \\
    ReSim~\cite{xiao2021region} & 200 & 66.1 & - & 39.8 & 60.2 & 43.5 & 36.0 & 57.1 & 38.6 & 41.4 & 61.9 & 45.4 & 37.5 & 59.1 & 40.3\\
    \rowcolor{Gray}
    LEWEL$_{M}$ & 200 & 68.1 & 88.6 & 40.0 & 59.8 & 43.7 & 36.1 & 57.0 & 38.7 & - & - & - & - & - & - \\
    \rowcolor{Gray}
    LEWEL$_{B}$ & 200 & \textbf{72.8} & \textbf{91.0} & \textbf{41.3} & \textbf{61.2} & \textbf{45.4} & \textbf{37.4} & \textbf{58.3} &\textbf{40.3} & \textbf{42.2} & \textbf{62.3} & \textbf{46.1} & \textbf{38.2} & \textbf{59.6} & \textbf{41.1} \\
    \midrule
    PixelPro~\cite{xie2021propagate} & 400 & 60.2 & 83.0 & 41.4 & 61.6 & 45.4 & 37.4 & - & - & - & - & - & - & - & - \\
    \rowcolor{Gray}
    LEWEL$_{B}$ & 400 & \textbf{73.8} & \textbf{91.7} & \textbf{41.9} & \textbf{62.4} & \textbf{46.0} & \textbf{37.9} & \textbf{59.3} & \textbf{40.7} & \textbf{43.4} & \textbf{63.5} & \textbf{47.7} & \textbf{39.1} & \textbf{60.7} & \textbf{42.4} \\
    \bottomrule
  \end{tabular}
  }
\end{table*}

\newpage
\appendix

\section{Additional Illustration of LEWEL}
\subsection{Illustration of the channel grouping operation.}
\label{sec:appendix_grouping}
\begin{figure}
    \centering
    \includegraphics[width=\linewidth]{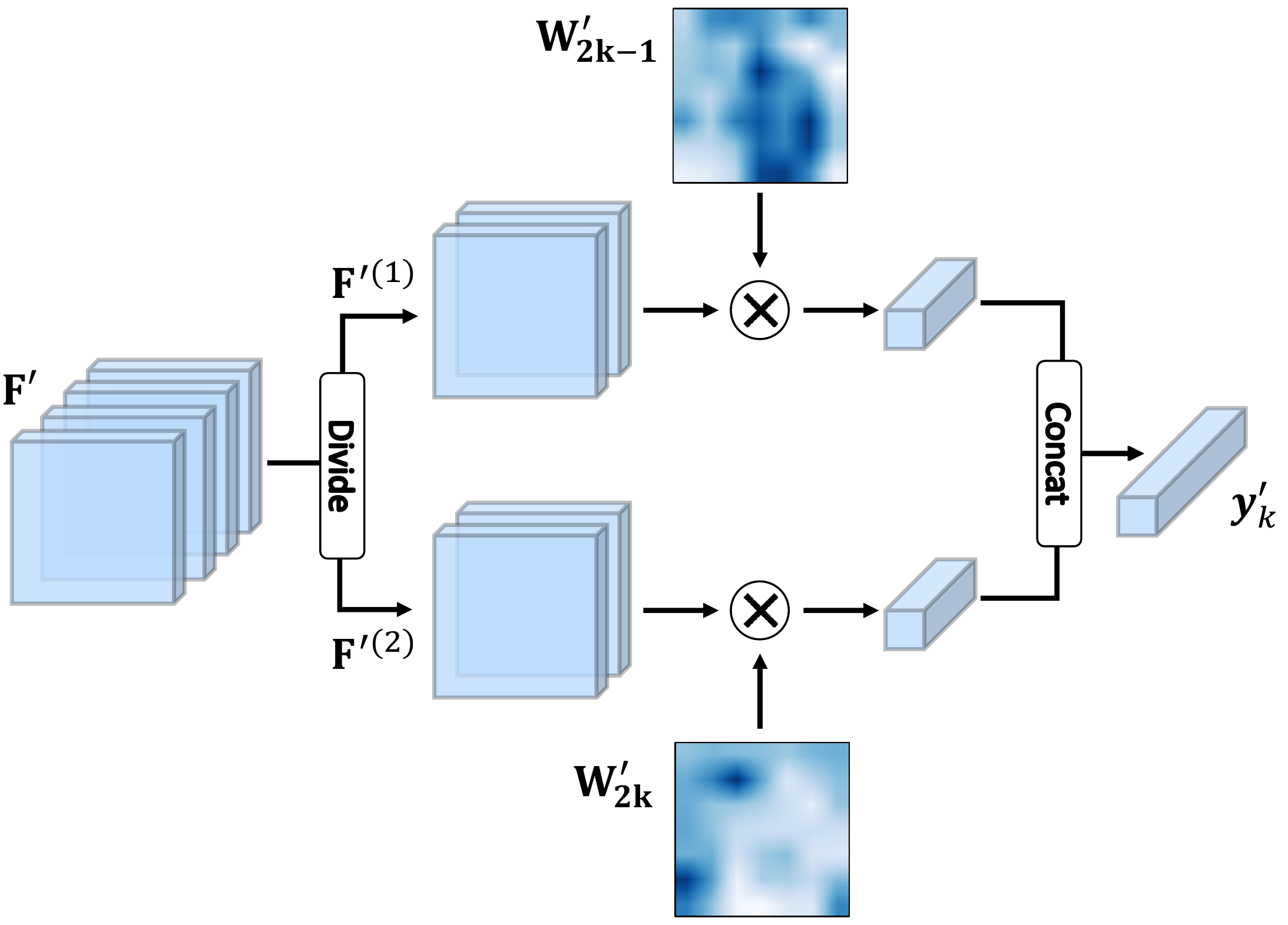}
    \caption{
    \textbf{An illustration of the channel grouping scheme}. Here the number of groups $h$ is set to $2$ for simplicity.
    }
    \label{fig:channel_grouping}
\end{figure}

Instead of using one alignment map to aggregate one aligned representation, we introduce a grouping scheme that divides the channels of $\mathbf{F}'$ uniformly into $h$ equal-size groups, that is $\mathbf{F}' = [\mathbf{F}'^{(1)}, \cdots, \mathbf{F}'^{(h)}]$ where $[\cdot]$ denotes the concatenation operation. 
Given the alignment maps $\{\mathbf{W}'_k\}_{k=1}^{d}$, we can accordingly aggregate a set of aligned representations $\{\y'_k : \y'_k \in \mathbb{R}^{D}\}_{k=1}^{d / h}$, and
{\small{
\begin{align}
    \y'_k = [\mathbf{W}'_{(k-1)\times h+1}\bm{\otimes}\mathbf{F}'^{(1)}, \cdots, \mathbf{W}'_{k\times h}\bm{\otimes}\mathbf{F}'^{(h)}], \forall k,
\end{align}}}
\hspace{-0.0575in}where $\bm{\otimes}$ is the spatial aggregation operation defined in \cref{eq:aggregation}. For a more intuitive illustration, we summarize the overview of this channel grouping scheme in \cref{fig:channel_grouping}, where we set the number of groups $h = 2$ for simplicity.

\begin{figure*}
    \centering
    \includegraphics[width=.95\linewidth]{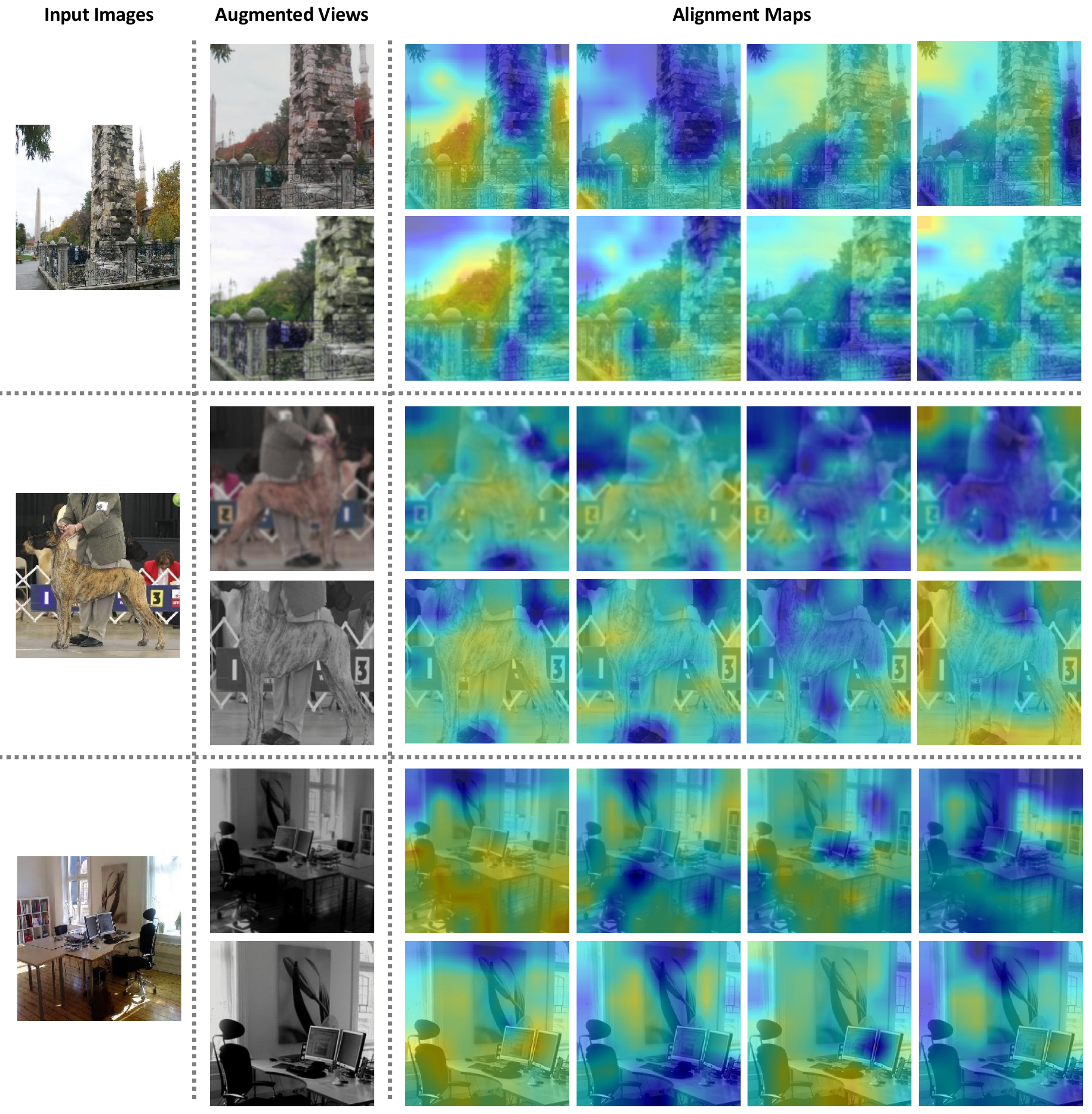}
    \caption{
    \textbf{Visualization of the alignment maps} predicted by LEWEL$_M$ on the ImageNet-1K validation set. First column: the input images from the ImageNet validation set. Second column: augmented views generated by random data augmentations. The rest columns: alignment maps predicted by LEWEL$_M$ based on augmented views. The visualization shows that LEWEL automatically finds semantically consistent alignments for self-supervised learning.
    }
    \label{fig:vis_alignment}
\end{figure*}

\section{Additional Experiment Results}

\subsection{Comparison with hand-crafted spatial alignment methods}
\label{sec:appendix_spt_ali}
Prior methods~\cite{wang2021dense,xie2021propagate,xiao2021region} are tailored for dense prediction and use pre-defined manual rules to match corresponding pixels. They emphasized on local feature learning and largely ignored the learning of global features that is also important in transferring to both classification and detection tasks (see Sec.~3.4 in~\cite{xie2021propagate}). In contrast, LEWEL is a generic method and benefits both image-level and dense predictions. LEWEL leverages the global projection head to predict the spatial alignment maps such that couples the learning of global features and aligned features.
Our experimental results in~\cref{tab:exp_spt_alm} show that LEWEL significantly outperforms~\cite{wang2021dense,xie2021propagate,xiao2021region} in terms of classification by up to 13\% while performing on par with or even better than~\cite{wang2021dense,xie2021propagate,xiao2021region} on detection/segmentation under $1\times$/$2\times$ training schedule, highlighting the generalization ability of LEWEL.

\section{Additional Analyses on LEWEL}
\subsection{Visualization of the alignment maps.}
\label{sec:appendix_vis}
In~\cref{fig:vis_alignment}, we visualize the alignment maps predicted by LEWEL$_M$ on the ImageNet-1K validation set, which suggest that LEWEL can automatically find semantically consistent alignments for self-supervised learning. Specifically, we observe that the alignment maps may activate on the region of an object (\eg, the visualization in the 3rd column, 1st-2nd rows), the region of multiple objects (\eg, the visualization in the 3rd column, 3rd-4th rows), and on the global region (\eg, the visualization in the 3rd column, 5th-6th rows). The visualization demonstrates that LEWEL is able to learn on both local and global representations simultaneously by manipulating the alignment maps.

\begin{table}
  \centering
  \small
  \caption{
  \textbf{Training time comparison} on with MoCov2 and BYOL on the ImageNet-1K dataset. The comparison is performed on a single machine with eight V100 GPUs using the automatic mixed precision (AMP) training in PyTorch 1.8.}
  \label{tab:training_time}
  \begin{tabular}{@{}lcc@{}}
    \toprule
    Method & Training Time/Epoch & Top1 Acc.@200 Epochs\\
    \midrule
    MoCov2 & 1213s & 64.5\\
    \rowcolor{Gray} LEWEL$_M$ & 1222s & 66.1\\
    \midrule
    BYOL & 1141s & 70.6\\
    \rowcolor{Gray} LEWEL$_B$ & 1191s & 71.5 \\
    \bottomrule
  \end{tabular}
\end{table}

\subsection{Computational cost.}
In~\cref{tab:training_time}, we compare the training time of LEWEL with that of the baselines. The comparison is performed on a single machine with eight V100 GPUs, CUDA 10.1, PyTorch 1.8, and the automatic mixed precision (AMP) training\footnote{
We find that AMP has little impact on the training time of MoCov2/LEWEL$_M$ but reduces that of BYOL/LEWEL$_B$ by $\sim$40\%.}.
For all methods, we report their training time of one epoch. According to the results in~\cref{tab:training_time}, we can observe that the training time of our method is only marginally increased compared to the baselines. To be more concrete, LEWEL$_B$ requires only $\sim$4\% additional overhead compared with BYOL while significantly outperforming BYOL, which demonstrates the efficiency of LEWEL.

\subsection{Limitations.}
One disadvantage of LEWEL is that the number of aligned representations largely depends on the output dimensionality $d$ of the coupled projection head $g$. In the cases where $d$ is very large, LEWEL might incur additional training overheads. Nevertheless, we note that our channel grouping scheme can mitigate this drawback by using a larger number of groups to reduce the number of aligned representations.